%% file: paper.tex
\documentclass[11pt]{article}

\usepackage{acl}

\usepackage{times}
\usepackage{latexsym}
\usepackage[T1]{fontenc}
\usepackage[utf8]{inputenc}
\usepackage{microtype}
\usepackage{inconsolata}

\usepackage{amsmath, amssymb}
\usepackage{booktabs}
\usepackage{tabularx}
\usepackage{graphicx}
\usepackage{enumitem}
\usepackage{hyperref}
\usepackage{float}   
\usepackage{algorithm}
\usepackage{algpseudocode}

\title{BlueprintAgent: Constraint-Triggered Targeted Revisits for\\
Simulation-Ready Generation from Scanned Structural Blueprints}

\author{
  \textbf{Zhouyuan Xu}, \textbf{Chen Yang}, \textbf{Linhao Wang},\\
  \textbf{Jiansheng Fan}, \textbf{Chen Wang}\\
  Tsinghua University, Beijing 100084, China\\
  \small{
    \textbf{Correspondence:}
    \href{mailto:fanjsh@tsinghua.edu.cn}{\texttt{fanjsh@tsinghua.edu.cn}},
    \href{mailto:chwang@tsinghua.edu.cn}{\texttt{chwang@tsinghua.edu.cn}}
  }
}

\hypersetup{
  pdftitle={BlueprintAgent: Constraint-Triggered Targeted Revisits for Simulation-Ready Generation from Scanned Structural Blueprints},
  pdfauthor={Zhouyuan Xu, Chen Yang, Linhao Wang, Jiansheng Fan, Chen Wang}
}

\begin{document}

\maketitle

\begin{abstract}
\input{sections/abstract}
\end{abstract}

\input{sections/introduction}
\input{sections/related_work}
\input{sections/method}
\input{sections/experiment}
\input{sections/conclusion}

\input{sections/limitations}

\section*{Ethical considerations}
\input{sections/ethics}

\section*{Acknowledgments}
\input{sections/acknowledgments}

\bibliography{refs}

\clearpage
\appendix
\input{sections/appendix}

\end{document}

%% file: sections/abstract.tex
\sloppy
Converting in-service reinforced-concrete (RC) building blueprints into simulation-ready models---structured frame representations that support deterministic FEM export and qualified-engineer review---underpins safety assessment and seismic retrofit, but the process remains manual. Direct prompting of a multimodal large language model (MLLM) over a scanned sheet is unreliable: outputs often violate engineering constraints on beam--column support, span count, or 3D continuity.
We present \textbf{BlueprintAgent (BPA)}, a constraint-triggered multimodal agent for simulation-ready frame extraction from scanned blueprints. BPA treats the MLLM as the primary reader and decision maker, with OCR and computer vision supplying localized evidence. Its central mechanism realizes engineering constraints as callable validators whose entity-level conflict reports trigger targeted MLLM revisits over the local region---an inference-time control distinct from fixed pipelines and free-form self-reflection.
We evaluate BPA on 300 real scanned blueprint sheets from 20 anonymized RC frame projects, against five baselines and six ablations. BPA reaches a macro-averaged Beam F1 of \textbf{0.994}, against 0.301 for single-MLLM zero-shot and 0.820 for a fixed pipeline; removing MLLM-led axis adjudication collapses Beam and Column F1 on complex multi-sheet projects. For dense technical drawings, engineering constraints are best deployed as triggers for entity-level targeted revisits rather than as post-hoc output filters.

%% file: sections/introduction.tex
\section{Introduction}
\label{sec:intro}

\begin{figure*}[t!]
\centering
\includegraphics[width=\textwidth,trim=0 22bp 0 11bp,clip]{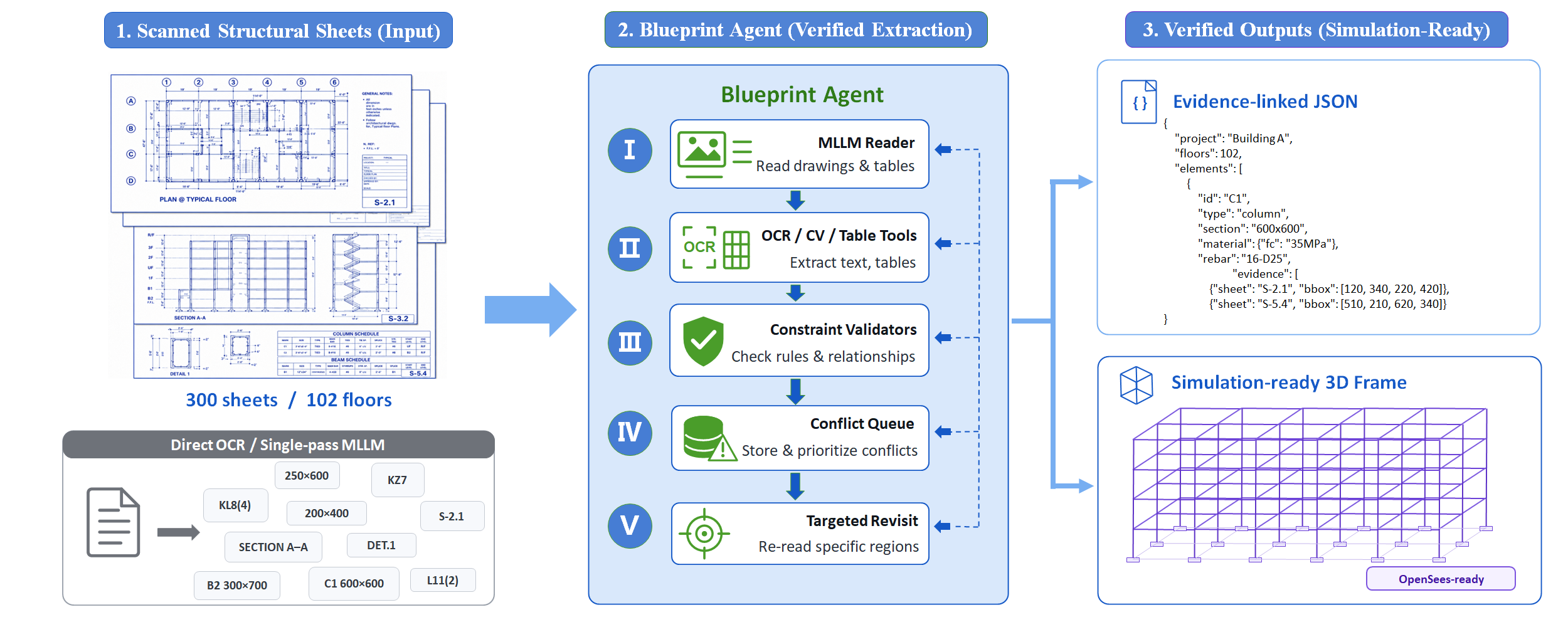}
\caption{BPA end-to-end pipeline: scanned sheets (left) are processed by a verified extraction pipeline (center) into evidence-linked JSON and a simulation-ready 3D frame (right).}
\label{fig:pipeline}
\end{figure*}

Multimodal large language models are increasingly applied to dense technical documents such as scientific tables, financial filings, and engineering drawings. The challenge in these tasks lies not in recognizing text or graphics on a page, but in producing an externally checkable, structurally consistent representation that downstream programs can consume. Scanned structural blueprints are a prototypical case: safety assessment, seismic retrofit, and digital-twin construction all require structural models, yet many in-service buildings exist only as paper or scanned construction drawings. Engineers read these axes, beams, columns, sections, elevations, materials, and notes across many sheets, then build FEM models by hand.

\textbf{Structural blueprints are not ordinary OCR targets.} The text they contain encodes engineering semantics, not free prose. For example, the \texttt{4} in \texttt{KL8(4)} denotes a four-span frame beam; \texttt{250$\times$600} becomes meaningful only when bound to a specific beam or column; an axis label is not merely a character but also defines a spatial coordinate in the simulatable 3D model. Detecting \emph{where} text or line segments appear is insufficient for producing a computable structural model. The system must know which label belongs to which beam, which beams are supported by which columns, which columns extend continuously across floors, and which dimensions can enter the FEM geometry.

Single-shot whole-sheet prompting of current MLLMs fails in three recurring ways that are generic to dense multimodal document understanding. Whole-sheet resolution downsampling drops small text, secondary axis labels, and dense dimension annotations. Without external constraint signals, the model over-connects: every grid intersection becomes a physical support and the system fabricates beam--column connections that are not actually there. Even schema-valid JSON can violate domain geometric, annotation, or 3D-consistency rules. These failure modes motivate a domain constraint layer that cross-checks perception output and re-reads local evidence at conflict sites.

We present \textbf{BlueprintAgent (BPA)}, a constraint-triggered multimodal reading-and-modeling agent. A project is represented as a hierarchy of \textbf{sheets, regions, evidence records, and a simulatable 3D model}. The MLLM acts as the primary perceiver and decision maker: it reads full sheets, main structural views, and local crops, and invokes OCR, CV-based axis probing, local re-OCR, axis-aware cropping, and table reading on demand. After each update, the system runs four families of constraints over the current simulatable 3D model: axis-grid, geometric-topological, annotation-consistency, and 3D-consistency. When a constraint fails, the system does not rewrite the model with hard rules; instead, it pushes a structured conflict item into a queue (for example, ``the endpoint \texttt{1-B} of beam \texttt{KL4} lacks column support''). The MLLM then revisits the relevant local region and, grounded in the evidence, decides to revise, mark as uncertain, or accept it as a documented anomaly.

This paper focuses on frame-level reconstruction within \textbf{RC frame structures}: axis grids, columns, primary beams, sections, floors, materials, 3D frame consistency, and simulation-ready output. Shear-wall, steel, and masonry structures, as well as slab reinforcement, column longitudinal reinforcement, stirrup densification zones, and joint-core reinforcement, are not main-task metrics.

Our contributions are:

\begin{enumerate}[leftmargin=*,topsep=2pt,itemsep=2pt]
\item \textbf{Method.} We propose BlueprintAgent, which couples MLLM-based reading with OCR / CV / local-re-OCR / axis-aware-cropping tools and four families of engineering-constraint validators. The key design is to let constraint failures trigger entity-level targeted revisits rather than act only as post-hoc filters. The validator interface is model-independent by design, and targeted replacement checks preserve the mechanism's improvement direction across GPT-5.4, Claude 4.6 Opus, and Qwen 3.6 Plus.
\item \textbf{Task.} We formalize \emph{generating simulation-ready structural models from scanned structural blueprints} as a multi-sheet technical-document understanding task. The input is the full set of scanned blueprint sheets per project; the output is evidence-linked structural JSON, a 3D frame, and OpenSees-ready FEM input.
\item \textbf{Evaluation.} We evaluate the design on \textbf{300 real scanned blueprint sheets covering 102 floors from 20 anonymized RC-frame projects}, reporting 5 baselines, 6 ablations, and 6 end-to-end metrics.
\end{enumerate}

%% file: sections/related_work.tex
\section{Related Work}
\label{sec:related}

\paragraph{Engineering drawing digitization.}
Research on engineering drawing digitization has explored object detection, OCR, image processing, and BIM reconstruction. Detector-based methods identify local targets such as beams, columns, and symbols \citep{zhao2020scanned_yolo}; BIM-oriented systems typically combine OCR, image processing, and hand-crafted rules to recover building components \citep{zhao2021bim_structural}. Recent MLLM-based work on engineering drawings attempts end-to-end parsing of annotations, GD\&T information, and local fields \citep{khan2024florence_engineering,khan2025drawings_decisions,khan2025multistage_hybrid}. Related work has also studied image-based structural analysis, capability-based evaluation of engineering agents, and the codification of expert knowledge for agent design \citep{altamirano2026photobeamsolver,molinari2026engiai,ulanuulu2026expert_knowledge}. These efforts demonstrate the feasibility of engineering drawing understanding but mostly focus on local elements or annotation extraction within a single drawing; closure of a multi-sheet simulatable 3D model has not yet been addressed. We instead target a multi-sheet simulatable 3D model whose correctness depends on cross-sheet text parsing, beam--column support topology, floor alignment, and FEM-readiness.

\paragraph{Floor plan and CAD understanding.}
Floor plan and CAD understanding centers on walls, doors, windows, rooms, and furniture, with representative datasets such as CubiCasa5K \citep{kalervo2019cubicasa5k} and FloorPlanCAD \citep{fan2021floorplancad}, floor-plan text detection and recognition \citep{schoenfelder2024floorplan_text}, and more recent MLLM-based floor-plan generation and editing \citep{qin2026housemind}. Structural blueprints differ from indoor floor plans: the core entities are axis grids, RC columns, frame beams, section labels, floor elevations, and material schedules; the target output is not a semantic segmentation map or room topology, but rather a load-bearing structural representation that engineering analysis can consume.

\paragraph{Visual document understanding and OCR.}
Visual document understanding models such as LayoutLMv3 \citep{huang2022layoutlmv3}, Donut \citep{kim2022donut}, and Pix2Struct \citep{lee2023pix2struct}, alongside multi-page document QA \citep{ma2024mmlongbench}, document-layout segmentation \citep{pfitzmann2022doclaynet}, logical document structuring \citep{li2024seg2act}, and recent LLM-based document understanding and verification \citep{zhu2025doclens,qian2025docrefine}, are isomorphic in output form to our ``drawing $\to$ structured JSON'' framing, but their structural boundary is defined by document elements rather than engineering entities under cross-sheet topological constraints. Scanned structural blueprints are harder: a single isolated number may be a primary-axis span, a sub-span, a local detailing dimension, or a table value, with semantics depending on its relation to the axis grid, the region, and surrounding components. BPA therefore consumes OCR and CV only as evidence under the active constraint context.

\paragraph{Tool-augmented and reflective agents.}
ReAct \citep{yao2023react}, Toolformer \citep{schick2023toolformer}, Reflexion \citep{shinn2023reflexion}, and Self-Refine \citep{madaan2023selfrefine} show that language models can interleave tool calls and improve outputs through feedback; Tree-of-Thoughts \citep{yao2023tot} extends this to deliberative search, while Chain-of-Verification \citep{dhuliawala2023chainverify} shows that external verifier signals can be used to reduce hallucinations. Recent document-oriented agents \citep{jin2025slideagent,li2024seg2act} and agent-evaluation benchmarks \citep{ye2024rotbench,bogin2024super} extend these ideas to multi-page documents, tool learning, and research repositories. BPA sits in this lineage but differs in the level of action and the trigger source: the verifier signal comes from a domain-level callable validator rather than a generic process reward model or free-form reflection, and constraints are injected into the MLLM's next planning step as structured JSON conflict items (containing entity ID, violated constraint, and supporting evidence), driving the MLLM to re-inspect the relevant local visual evidence.

\paragraph{Constrained and structured generation.}
Constrained generation in NLP includes lexically constrained decoding \citep{hokamp2017grid,post2018dba}, grammar-constrained decoding \citep{geng2023grammar}, and neuro-symbolic decoding \citep{lu2022neurologic}. Planning--verification loops have also been applied to factuality and logical consistency checking \citep{dhuliawala2023chainverify}. BPA shares with these efforts the idea of \emph{constraints participating in generation}, but operates at a different level: token-level constrained decoding regulates a single text generation pass, and natural-language neuro-symbolic methods act on textual inference; BPA consumes constraint-failure signals to re-acquire visual evidence, re-extract entities, and update the simulatable 3D model.

\paragraph{Feature comparison with prior work.}
Appendix~\ref{sec:appendix_rw} contrasts BPA with representative verification, reflection, and tool-augmented agents along four key design dimensions (Table~\ref{tab:rw}).

\paragraph{Positioning of this work.}
This work sits at the intersection of engineering drawing digitization, visual document understanding, tool-augmented multimodal agents, and constraint-aware structured generation. Our contribution is not a new OCR model, detector, or FEM solver, but a reading-perception design built on an MLLM for dense technical drawings: it converts domain constraints into auditable triggers for multimodal evidence collection and entity revision.

%% file: sections/method.tex
\section{Task and Data}
\label{sec:task}

\paragraph{Input unit.}
The input unit is \textbf{a complete set of scanned structural blueprints for one RC frame project}: a project $S = \{s_1, \ldots, s_n\}$ typically contains from a few to several dozen sheets, covering beam, column, and slab plan views, title blocks, material notes, story-height tables, and detail views. The agent aggregates cross-sheet evidence and makes structural decisions over it. Each sheet typically holds one or two primary structural views (e.g., beam, column, or slab plans; elevations; sections), plus auxiliary regions such as joint or stairway details. A sheet is further partitioned into multiple regions $s_i \to \{r_i^{\textrm{main}}, r_i^{\textrm{table}}, r_i^{\textrm{title}}, r_i^{\textrm{note}}, r_i^{\textrm{detail}}, \ldots\}$. The system labels each region with a role but tolerates label errors; the role only governs how the region's evidence enters the simulatable 3D model.

\paragraph{Output.}
The canonical output is an evidence-linked structured JSON state $H$, comprising: an axis grid (labels, ordering, span lengths, grid intersections); floors (labels, elevations, story heights, corresponding sheets); columns (object ID, axis-grid position, section, floor, status, confidence, evidence references); primary beams (beam identifier, node sequence, span count, section, orientation, status, confidence, evidence references); materials (concrete grade, rebar grade, or other material specifications); and validation status (constraint check results, uncertain items, accepted anomalies). The MLLM creates and revises $H$ from localized evidence. Once the key geometric and material fields are complete, a deterministic exporter generates an OpenSees \texttt{.tcl} file, and the 3D frame visualizes the same state.

\paragraph{Scope.}
This paper addresses frame-level structural reconstruction only: axes, columns, primary beams, sections, floors, materials, and 3D frame consistency. Fine-grained reinforcement (slab positive and negative reinforcement, column longitudinal reinforcement, stirrup densification zones, joint-core reinforcement) does not enter the main experimental metrics. Such reinforcement may be recorded as \texttt{non-core region} content or auxiliary evidence, but it is not used to evaluate simulation-ready frame recovery.

\paragraph{Dataset.}
The evaluation set uses \textbf{sheet} as the primary unit of count: \textbf{300 real scanned structural blueprint sheets} covering \textbf{102 floors}, drawn from partner-authorized structural archives for 20 anonymized RC frame projects, spanning beam, column, and slab plan views, foundation plans, elevations, sections, title blocks, material notes, story-height tables, and detail views. GPT-5.4 is used for inference without parameter fine-tuning, and the 300 sheets form the evaluation corpus. Difficulty tiers are split into simple, medium, and complex based on the number of floors, per-project sheet count, axis-grid density, and scan quality.

\paragraph{Ground truth.}
Complete project-level ground-truth JSON and OpenSees \texttt{.tcl} models were manually annotated from the full project sheet sets and reviewed entity by entity under the structural annotation protocol.

\paragraph{Data availability.}
The public academic release accompanying this paper contains the 300 anonymized blueprint sheets, project-level ground-truth JSON, manually annotated OpenSees \texttt{.tcl} models, a project manifest, the ground-truth schema and annotation protocol, evaluation code, and a datasheet.

\paragraph{Terminology.}
Throughout the paper we use \emph{sheet} (one scanned blueprint), \emph{region} (a semantic region inside a sheet), \emph{entity} (a structural object: axis / column / beam / floor / material), \emph{evidence} $E$ (a localized OCR / CV / MLLM observation), \emph{conflict} $q \in Q$ (a validator violation report), and \emph{simulatable 3D model} $H$ (the project-level evidence-linked structural representation). A \emph{validator} is a callable constraint checker $v: H \to \textrm{List}[\textrm{Conflict}]$; a \emph{targeted revisit} is the MLLM action of revisiting relevant local evidence and revising an entity, triggered by a validator conflict signal. \texttt{KL} / \texttt{KZ} are the frame-beam / frame-column labeling prefixes in Chinese structural drawings (e.g., \texttt{KL8(4)} = frame beam No. 8, four-span); the \texttt{axis\_px\_span\_ratio} gate (Appendix~\ref{sec:appendix_terminology}) checks axis-span pixel ratios against annotated span ratios within $\pm15\%$ and corresponds to ablation A5.

\begin{figure*}[t]
\centering
\includegraphics[width=\textwidth,trim=0 76bp 0 30bp,clip]{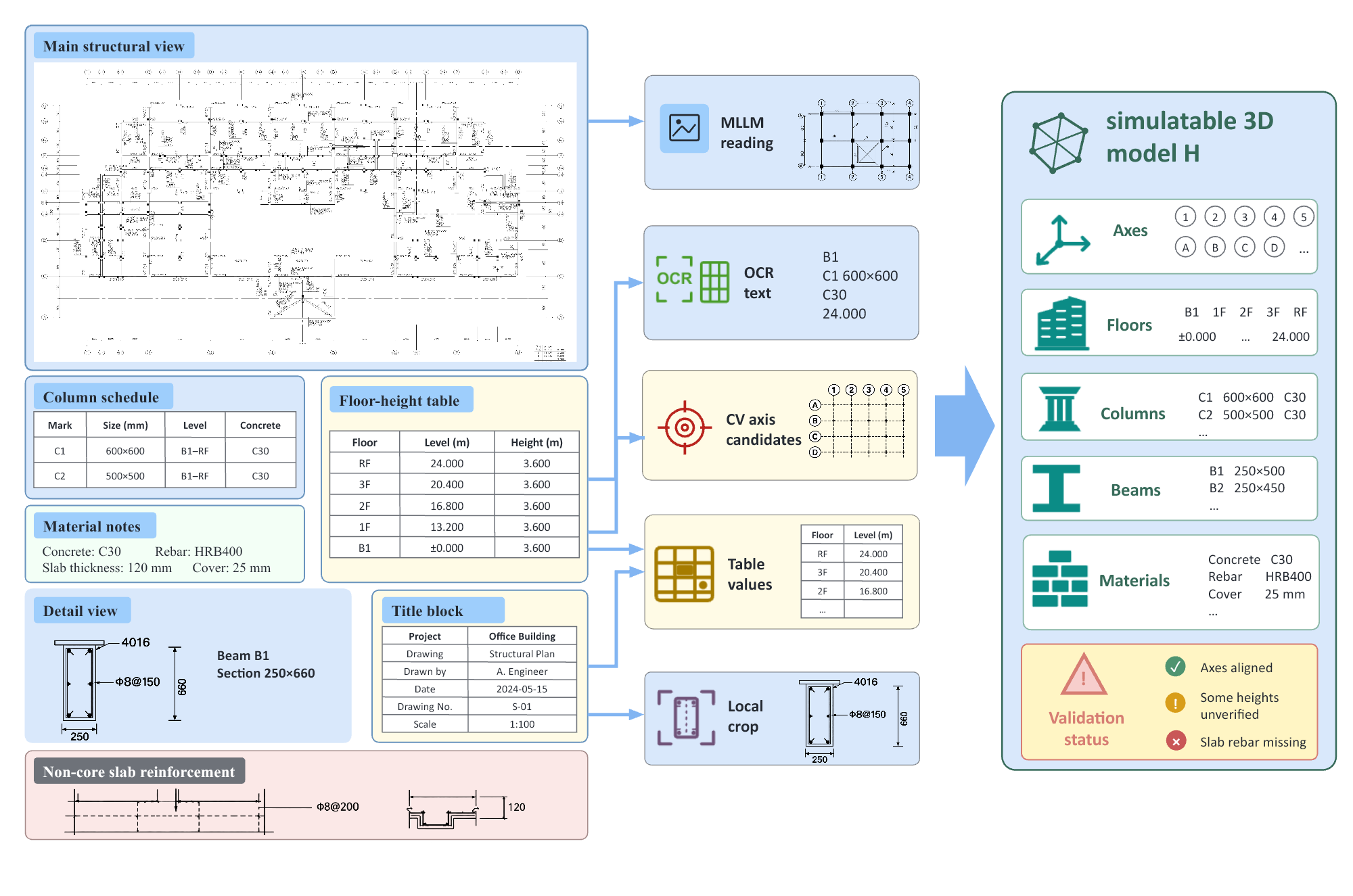}
\caption{Sheet--region--evidence representation: a sheet (left) is partitioned into regions whose evidence (center) flows into the simulatable 3D model with a validation-status panel (right).}
\label{fig:sre}
\end{figure*}

\section{Method}
\label{sec:method}

\paragraph{Overview.}
BPA is a tool-augmented multimodal agent (Figure~\ref{fig:pipeline}; full pseudocode in Appendix~\ref{sec:appendix_algorithm}). It builds a project-level evidence pool through OCR caching and sheet--region analysis, then sequentially handles the core entity classes (axis grid, columns, beams, floors, materials) through axis-aware cropping, primary-element extraction, semantic validation, 3D consistency check, and FEM export, gradually closing the simulatable 3D model $H$. Within each stage, an \emph{assess--plan--execute--interpret--update--validate--revisit} loop consumes validator feedback and performs targeted revisits on low-confidence or conflicting entities until convergence or budget exhaustion.

\paragraph{Design choice: MLLM as the primary reader.}
Real scanned blueprints vary widely in line work, annotation style, resolution, and aging; the key problems are not local detection (where task-specialized detectors like YOLO or DETR excel) but cross-sheet aggregation, 3D topology alignment, and engineering-exception judgment. BPA therefore takes the MLLM as the primary reader and decision maker; OCR, CV, and table tools provide evidence. The MLLM's three roles map to specific ablation cells: perception to B1/A1/A2; reasoning to A6 (axis adjudication); decision to B4 (constraint feedback disabled during inference).

\paragraph{Sheet--region--evidence simulatable 3D model.}
BPA maintains the simulatable 3D model
\begin{multline*}
H = \{\textrm{sheets}, \textrm{regions}, \textrm{axes}, \textrm{floors}, \\
\textrm{columns}, \textrm{beams}, \textrm{materials}, E, V, Q\},
\end{multline*}
where $E$ is the set of evidence records, $V$ the validation state, and $Q$ the conflict queue. Each entity carries \texttt{status}, \texttt{confidence}, \texttt{evidence\_refs}, and \texttt{provenance}. Structural reasoning is grounded in axis coordinates, node sequences, real dimension annotations, floors, and component attributes, not in pixel distances. Figure~\ref{fig:sre} shows how a sheet is partitioned into regions and how evidence flows into $H$.

\paragraph{Evidence tools.}
The tool library splits into \emph{perceivers}---MLLM reading, multi-scale OCR cache, local re-OCR (multi-candidate readings inside an MLLM-specified box), and a CV axis probe (axis-label circles, lines, local symbols)---and \emph{methods}---layout region detection, axis-aware cropping (preserving axis-label context), beam / column / text-binding candidate generation, the four semantic validators below, and an FEM exporter to OpenSees-ready \texttt{.tcl}.

\paragraph{Axis-grid-aware spatial representation.}
The axis grid is the spatial skeleton of the simulatable 3D model. Columns and beams are represented by grid intersections, axis labels, and node sequences, not by pixel coordinates. Axis-label adjudication follows two principles. \emph{MLLM first:} when an axis label is visually readable, the MLLM reads and adjudicates it directly. \emph{On-demand evidence:} when an axis label, secondary axis, or span length is unclear, the agent requests local OCR, CV axis candidates, or a region crop. As a design-time safety gate, the system additionally maintains an \texttt{axis\_px\_span\_ratio} sanity check, which compares the pixel-distance ratio between adjacent axis spans against the annotated span ratio (tolerance $\pm15$\%); a significant deviation flags an axis-grid misalignment and triggers a targeted revisit. The gate targets a low-frequency but high-impact failure mode (axis-grid misalignment) and its ablation corresponds to A5.

\begin{figure*}[t]
\centering
\includegraphics[width=0.96\textwidth]{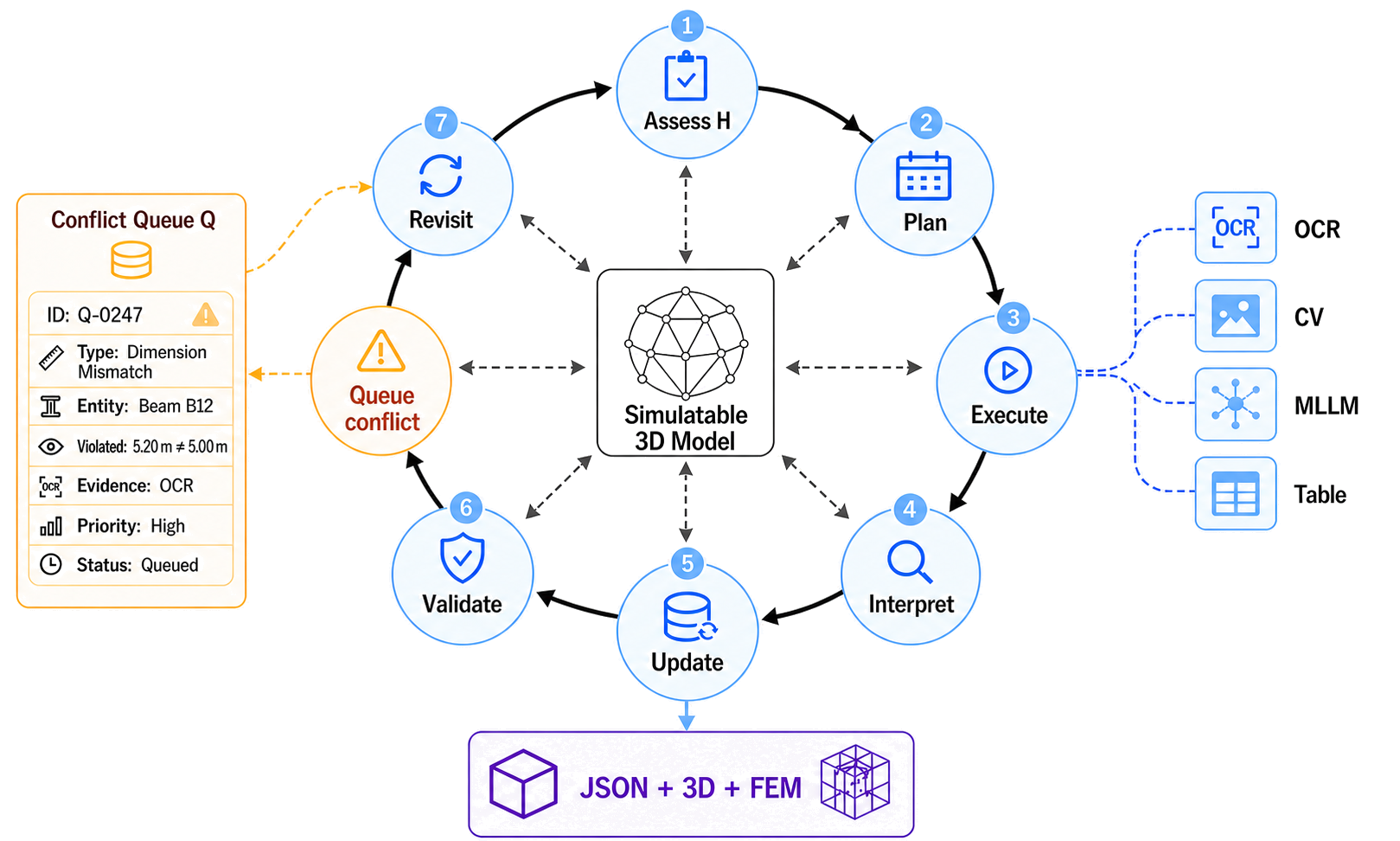}
\caption{Targeted revisit loop: a seven-step inner loop consumes evidence from four tools and uses a structured conflict record (left) to revise the simulatable 3D model.}
\label{fig:revisit}
\end{figure*}

\paragraph{Targeted revisit mechanism.}
The core mechanism of BPA is \textbf{validator-triggered targeted revisit}, an inference-time control mechanism in which domain-level validators emit structured conflict signals that gate entity-level multimodal revisitation. When a validator detects a conflict, the system injects the structured conflict item $q \in Q$ as evidence into the next \texttt{plan} call, letting the MLLM revisit the relevant region with the specific conflict in hand. This loop resembles ReAct's thought-action cycle, Reflexion's self-reflection signal, and Self-Refine's critique-revise. Its \textbf{trigger source} is an external schema-level constraint rather than internal natural-language reflection, distinguishing verification-driven from self-reflection-driven agents (see Section~\ref{sec:related} and Table~\ref{tab:rw}). BPA converts errors into auditable conflict records and steers the MLLM back to the corresponding visual evidence. Figure~\ref{fig:revisit} illustrates the inner loop with a representative conflict record.

\paragraph{Constraint validators.}
BPA exposes four families of validators. $C_{\textrm{axis}}$ (axis-grid constraints) checks axis-label legality, ordering, span-length evidence, secondary axes, and cross-floor axis-grid consistency; it is a spatial prerequisite and is not removed as an isolated ablation. $C_{\textrm{geo}}$ (geometric-topological constraints) checks whether columns sit at grid intersections, whether primary beams run along grid lines, whether beam endpoints rest on columns or legitimate supports, and whether continuous beams pass through intermediate supports. $C_{\textrm{anno}}$ (annotation-consistency constraints) checks whether the span count in \texttt{KL8(4)} is consistent with topology, whether \texttt{250$\times$600} is bound to the correct beam or column, and whether KZ identifiers can be matched to the column schedule or column-annotation evidence. $C_{\textrm{3D}}$ (3D consistency constraints) checks column continuity across floors, beam support, floor-elevation monotonicity, and component attribute completeness.

Validators only emit lists of violations; corrective actions are taken by the MLLM during the revisit loop. Formally, each $v \in C$ is a mapping $v: H \to \textrm{List}[\textrm{Conflict}]$, and $\textsc{validate}(H, C) = \bigcup_{v \in C} v(H)$. The current implementation defines 12 conflict types across $C_{\textrm{geo}}$, $C_{\textrm{anno}}$, and $C_{\textrm{3D}}$ ($C_{\textrm{axis}}$ is a spatial prerequisite and is not exposed as a separate ablation; full type list in the supplementary material). Load constraints and code-compliance constraints (minimum reinforcement ratios, seismic category) are left to future work.

\paragraph{Conflict handling.}
When a constraint fails, the validator writes a structured record into $Q$ with fields \texttt{type / entity / violated / evidence / priority / status} (an example record appears in Figure~\ref{fig:revisit}; numerical checks may add \texttt{observed} / \texttt{expected} / \texttt{severity}). In the revisit, the MLLM takes one of three actions: \emph{revise} (re-read local evidence and modify the entity), \emph{mark uncertain} (flag for human review), or \emph{accept anomaly} (accept a legitimate exception, e.g., cantilever beam, intentionally terminated column). This avoids forcing brittle rules onto the many legitimate engineering exceptions that real structural drawings contain.

%% file: sections/experiment.tex
\section{Experiments}
\label{sec:experiment}

\subsection{Research questions}
Our experiments answer three questions. \textbf{RQ1 (Effectiveness):} does treating engineering constraints as entity-level targeted-revisit triggers improve end-to-end structural extraction? \textbf{RQ2 (Mechanism attribution):} which components are most critical on complex multi-sheet projects (cross-sheet aggregation A3, $C_{\textrm{3D}}$ A4, the \texttt{axis\_px\_span\_ratio} gate A5, MLLM-led axis adjudication A6)? \textbf{RQ3 (Bounds):} where do BPA's gains over pure-OCR rules, single-MLLM zero-shot, and the fixed pipeline come from?

\subsection{Baselines}
Five systems are compared, all MLLM-driven ones sharing the same backbone so that gains are attributable to framework design.
\textbf{B1: Pure-OCR rules}---only OCR and hand-written rules; no MLLM.
\textbf{B2: Single-MLLM zero-shot}---feeds a full sheet to the MLLM for direct structural JSON; no tools, constraints, or revisits.
\textbf{B3: Fixed pipeline}---invokes OCR, CV, layout, and candidate-generation tools in a fixed order; no agent decision making, engineering-constraint validation, or conflict queue.
\textbf{B4: Agent without constraint feedback}---retains the agent loop, tool calls, and output contract, but does not feed conflicts from $C_{\textrm{geo}}$, $C_{\textrm{anno}}$, or $C_{\textrm{3D}}$ back into inference to trigger revisits.
\textbf{B5: Full BlueprintAgent}---the complete system including the four constraint-validator families, the conflict queue, and targeted revisits.

\subsection{Ablations}
Six ablations each remove one component from B5.
\textbf{A1 (no axis-aware cropping):} naive bounding-box crops only.
\textbf{A2 (no axis-label repaste):} local crops lose global axis-label context.
\textbf{A3 (no cross-sheet aggregation):} cross-sheet structures merged only by naive ID matching.
\textbf{A4 (no $C_{\textrm{3D}}$ closure):} removes the cross-floor 3D-closure stage from B5.
\textbf{A5 (no \texttt{axis\_px\_span\_ratio} gate):} removes the consistency check between axis-span pixel ratios and annotated span lengths.
\textbf{A6 (no MLLM-led axis adjudication):} accepts CV axis candidates directly, without MLLM re-checking.

\subsection{Metrics}
The evaluation suite contains six end-to-end metrics (full definitions in Appendix~\ref{sec:appendix_metrics}): \textbf{Axis F1} (axis-label match + position IoU $\geq 0.85$), \textbf{Col F1} (column ID match after grid alignment), \textbf{Beam F1} (beam ID match + endpoint axis multiset match), \textbf{Bind F1} (KL text content + target beam ID match), \textbf{JSON Validity} (output passes the schema), and \textbf{Structured-Model Completeness (SMC)}. The project-level SMC gate equals 1 iff $C_{\textrm{3D}}$ validation is enabled, the normalized evaluator status is \texttt{OK}, the predicted column and primary-beam inventories are both nonempty, and each has at least one ground-truth match; otherwise it equals 0. We report the project-equal mean over all 20 projects. Every variant uses the same output contract and evaluator, except that A4 disables the $C_{\textrm{3D}}$ gate by definition.

\paragraph{Reproducibility setup.}
Each main-table cell reports one complete run. Except for deterministic B1, all systems call GPT-5.4 at temperature 0.2 with a fixed system prompt; per-cell prompts, runtime configurations, and metric JSON files are provided in the supplementary material. Run-to-run stability for the mechanism-bearing B4--B5 contrast and controlled backbone replacements are evaluated on a prespecified stratified five-project subset using project-equal aggregation. The main tables use the 20-project evaluation.

\paragraph{Metric behavior across variants.}
Axis F1, Col F1, Bind F1, JSON Validity, and SMC reach 1.000 for B5 while measuring distinct properties: axis identity and position, post-aggregation column inventory, KL-to-beam binding, schema conformance, and protocol-level structured-model completion. Each metric decreases in at least one baseline or ablation. The B3, B4, and A6 cells identify the mechanism-bearing gaps, while the remaining 0.6\% Beam-F1 error for B5 is concentrated in the perception and dense-binding cases reported in Appendix~\ref{sec:appendix_errors}.

\subsection{Main results}
\label{sec:main_results}
Table~\ref{tab:main} presents the main results. \textbf{Perception-only systems cannot recover structural topology.} B1's Axis F1 and Beam F1 are both 0.000 (schema-valid but empty JSON); B2's Axis F1 reaches 0.987 but its Beam F1 is 0.301 and Col F1 is 0.537. Whole-sheet prompting can identify part of the axis grid on simple drawings but cannot jointly verify span count, supporting columns, and beam topology in one inference pass. \textbf{The fixed pipeline improves perception but cannot self-correct.} B3 adds OCR, CV, and candidate generation, raising Beam F1 to 0.820---still well below B5's 0.994: once an early axis-label, candidate, or binding error occurs, a fixed pipeline has no mechanism to localize the error back to the visual evidence and re-read it.

\textbf{Constraint-triggered revisits drive the main gain.} B4 retains the agent loop and tools but suppresses engineering-constraint feedback during inference; its Beam F1 is close to B3, whereas B5 improves Col F1, Beam F1, and Bind F1. JSON Validity and SMC are already saturated for B4. The B4--B5 contrast therefore localizes the gain to integrating structured conflicts into the loop as entity-level revisit triggers.

\begin{table}[t]
\centering
\scriptsize
\setlength{\tabcolsep}{3pt}
\begin{tabular*}{\columnwidth}{@{\extracolsep{\fill}}lcccccc@{}}
\toprule
\textbf{Method} & \textbf{Axis} & \textbf{Col} & \textbf{Beam} & \textbf{Bind} & \textbf{JSON} & \textbf{SMC} \\
\midrule
B1: OCR rules        & 0.000 & 0.000 & 0.000 & 0.000 & 1.000 & 0.000 \\
B2: MLLM zero-shot   & 0.987 & 0.537 & 0.301 & 0.327 & 1.000 & 0.350 \\
B3: Fixed pipeline   & 0.991 & 0.920 & 0.820 & 0.868 & 1.000 & 1.000 \\
B4: no $C$ feedback & 0.993 & 0.959 & 0.768 & 0.873 & 1.000 & 1.000 \\
\textbf{B5: Full BPA} & \textbf{1.000} & \textbf{1.000} & \textbf{0.994} & \textbf{1.000} & \textbf{1.000} & \textbf{1.000} \\
\bottomrule
\end{tabular*}
\caption{Main results (300 sheets / 102 floors; single run; macro average over 20 projects). Column headers abbreviate Axis F1, Col F1, Beam F1, KL Bind F1, JSON Schema Validity, and Structured-Model Completeness. B4 suppresses $C_{\textrm{geo}}$/$C_{\textrm{anno}}$/$C_{\textrm{3D}}$ feedback during inference while retaining the common output contract and evaluator.}
\label{tab:main}
\end{table}

\subsection{Ablation study}
Table~\ref{tab:ablation} reports the six ablations. A6 (no MLLM-led axis adjudication) has the largest extraction decline: macro Beam F1 drops to 0.369 and Col F1 to 0.614, with the effect intensifying on multi-floor multi-sheet composite-grid projects where an erroneous axis grid corrupts both the column inventory and beam topology. A3 (no cross-sheet aggregation) most strongly affects column deduplication: the same physical column is counted across plan views, inflating the inventory by 2.0--3.0$\times$ and reducing Col F1 to 0.516. A5 (no \texttt{axis\_px\_span\_ratio} gate) yields Beam F1 of 0.828 and Col F1 of 0.831, consistent with its focus on infrequent axis-grid misalignment. A4 retains planar extraction metrics but disables $C_{\textrm{3D}}$ validation, so the protocol assigns SMC 0 to every project.

\begin{table}[t]
\centering
\scriptsize
\setlength{\tabcolsep}{3pt}
\begin{tabular*}{\columnwidth}{@{\extracolsep{\fill}}lcccc@{}}
\toprule
\textbf{Variant} & \textbf{Axis} & \textbf{Col} & \textbf{Beam} & \textbf{SMC} \\
\midrule
B5 Full BPA              & 1.000 & 1.000 & 0.994 & 1.000 \\
A1: no crop              & 1.000 & 0.838 & 0.674 & 0.900 \\
A2: no repaste           & 1.000 & 0.827 & 0.896 & 0.850 \\
A3: no x-sheet aggr.\    & 0.981 & 0.516 & 0.564 & 0.850 \\
A4: no $C_{\textrm{3D}}$ closure & 1.000 & 0.648 & 0.551 & 0.000 \\
A5: no span-ratio gate   & 1.000 & 0.831 & 0.828 & 0.850 \\
A6: no axis adjud.\      & 0.892 & 0.614 & 0.369 & 0.650 \\
\bottomrule
\end{tabular*}
\caption{Ablation results. A1 = no axis-aware crop; A2 = no axis-label repaste; A3 = no cross-sheet aggregation; A4 = no $C_{\textrm{3D}}$ closure; A5 = no \texttt{axis\_px\_span\_ratio} gate; A6 = no MLLM-led axis adjudication.}
\label{tab:ablation}
\end{table}

\begin{figure*}[t!]
\centering
\IfFileExists{figures/figure4-final.pdf}{%
  \includegraphics[width=\textwidth]{figures/figure4-final.pdf}%
}{%
  \IfFileExists{figures/figure4-final.png}{%
    \includegraphics[width=\textwidth]{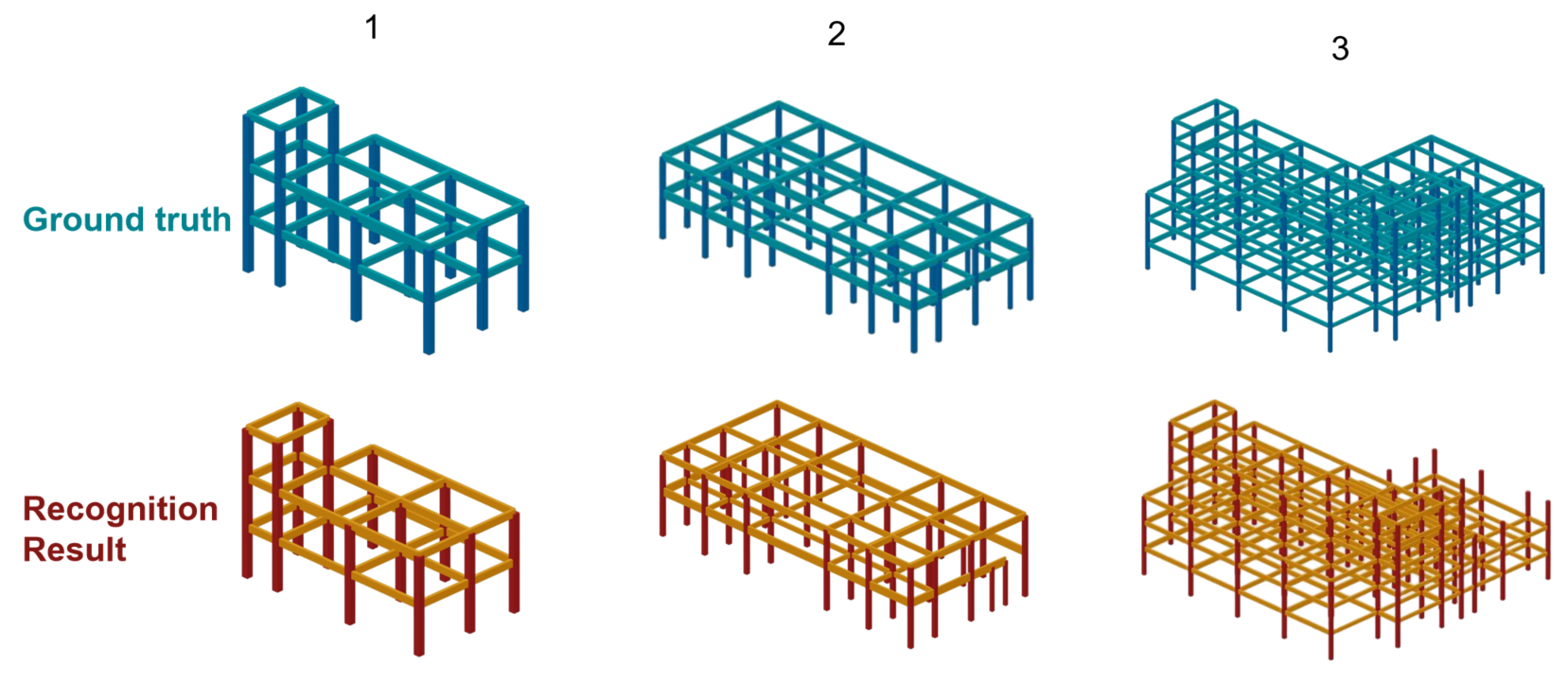}%
  }{%
    \begingroup
    \setlength{\tabcolsep}{1.5pt}
    \setlength{\fboxsep}{0pt}
    \begin{tabular}{ccc}
      \fbox{\parbox[c][0.19\textheight][c]{0.305\textwidth}{\centering (a)}} &
      \fbox{\parbox[c][0.19\textheight][c]{0.305\textwidth}{\centering (b)}} &
      \fbox{\parbox[c][0.19\textheight][c]{0.305\textwidth}{\centering (c)}} \\[2pt]
      \fbox{\parbox[c][0.19\textheight][c]{0.305\textwidth}{\centering (d)}} &
      \fbox{\parbox[c][0.19\textheight][c]{0.305\textwidth}{\centering (e)}} &
      \fbox{\parbox[c][0.19\textheight][c]{0.305\textwidth}{\centering (f)}}
    \end{tabular}
    \endgroup
  }%
}
\caption{Representative 3D frame modeling results for three projects. Each column pairs the ground-truth model (top) with the corresponding BlueprintAgent output (bottom) under the same orthographic camera; colors distinguish columns and beams.}
\label{fig:modeling_results}
\end{figure*}

\subsection{Targeted robustness and isolation checks}
\label{sec:targeted_checks}
Table~\ref{tab:targeted_checks} summarizes targeted checks on a prespecified stratified subset containing all three complex projects and one medium and one simple project. For GPT-5.4, three repeats per B4/B5 configuration give project-equal Beam F1 of $0.7894\pm0.0121$ for B4 and $0.9609\pm0.0074$ for B5, where the dispersion is pooled within-project run SD; all 15 project-by-repeat-index contrasts favor B5. Applying the same validators only after B4 terminates yields B6 Beam precision/recall/F1 of $0.8996/0.8069/0.8505$, compared with $0.9518/0.9704/0.9609$ for B5. Thus B5 exceeds B6 by 0.1104 Beam F1, and 23 of 32 conflicts classified as requiring localized visual rereading are resolved after rereading. Controlled replacements with Claude 4.6 Opus and Qwen 3.6 Plus preserve the B5-over-B4 direction on every evaluated project. Together, these checks support both the run stability of the central contrast and the contribution of evidence reacquisition beyond post-hoc constraint processing.

\begin{table}[t]
\centering
\footnotesize
\setlength{\tabcolsep}{2.5pt}
\begin{tabular}{@{}lccc@{}}
\toprule
\textbf{Backbone} & \textbf{B4} & \textbf{B6} & \textbf{B5} \\
\midrule
GPT-5.4 & $0.7894{\pm}0.0121$ & 0.8505 & $0.9609{\pm}0.0074$ \\
Claude 4.6 Opus & 0.7846 & -- & 0.9062 \\
Qwen 3.6 Plus & 0.7787 & -- & 0.8829 \\
\bottomrule
\end{tabular}
\caption{Project-equal Beam F1 on the targeted five-project subset. GPT-5.4 reports three repeats per B4/B5 configuration; B6 is deterministically derived from the corresponding B4 states. Replacement-backbone checks use one B4/B5 run per project.}
\label{tab:targeted_checks}
\end{table}

\subsection{Case studies and counterexamples}
\label{sec:cases}
Figure~\ref{fig:modeling_results} presents three representative modeling cases; Appendix~\ref{sec:appendix_cases} records their diagnostic mechanisms. The A6 counterexample admits \emph{ghost axes} when CV candidates bypass MLLM re-checking, corrupting the column inventory and beam topology. A second case passes $C_{\textrm{3D}}$ and runs in OpenSees but yields anomalous top-floor displacement, reinforcing qualified-engineer review before engineering use.

\subsection{Findings and analysis}
The small B3--B4 and large B4--B5 gaps identify structured conflicts consumed as entity-level revisit triggers, rather than looping alone, as the operative mechanism. B6 confirms that post-hoc constraints do not recover the recall gained through localized rereading. The consistent B5-over-B4 direction across three tested backbones supports interface portability, with larger gains on complex projects (Appendix~\ref{sec:appendix_pertier}).

\subsection{Adjudicated consensus error profile}
Two structural engineers independently categorized residual B5 errors, with disagreements arbitrated by a third (Appendix~\ref{sec:appendix_errors}). The adjudicated profile is 45\% perception, 25\% topology, 20\% binding, and 10\% completeness: perception dominates, while the remaining categories concentrate in dense regions or structured-model closure.

%% file: sections/conclusion.tex
\section{Conclusion}
\label{sec:conclusion}

BlueprintAgent, the system introduced in this paper, treats engineering constraints not as a final-stage filter but as triggers that pull the MLLM back to the specific local region where its first read went wrong. The hierarchy is sheet $\to$ region $\to$ evidence $\to$ simulatable 3D model; the validators are callable, schema-level, and entity-scoped. On 300 real scanned structural blueprint sheets covering 102 floors, this pattern consistently improves beam topology, column identification, and text-binding metrics over single-MLLM zero-shot and over a fixed OCR-plus-CV pipeline that lacks the revisit mechanism.

%% file: sections/limitations.tex
\section*{Limitations}
\label{sec:limitations}

\paragraph{Sample scope.}
Our main evaluation uses 300 sheets covering 102 floors from 20 anonymized RC-frame projects. The empirical scope is RC frames; shear-wall, steel, masonry, and complex hybrid structures remain outside the evaluation.

\paragraph{Annotation depth and scale trade-off.}
Each core entity (axis, column, beam, KL label, floor, section) requires structural-engineer-level annotation and review. The evaluation prioritizes structural-semantic annotation depth across 300 sheets. Scaling to a larger drawing library requires corresponding engineering-annotation effort.

\paragraph{Backbone scope.}
The main comparison fixes GPT-5.4 as the perception-and-decision core to avoid a model confound. Controlled B4/B5 replacements with Claude 4.6 Opus and Qwen 3.6 Plus preserve the same improvement direction on the targeted five-project subset. Our empirical portability claim is limited to these three tested backbones.

\paragraph{Evaluation repetition.}
Main results and ablations contain one complete run per cell (except deterministic B1). The targeted stability check therefore repeats the mechanism-bearing B4--B5 comparison three times per configuration on a prespecified stratified five-project subset. This check establishes stability for the central contrast, while the complete main matrix remains a single-run evaluation.

\paragraph{Single FEM solver.}
We export to OpenSees and verify that linear-elastic static analysis runs. Dynamic analysis, push-over, time-history, and nonlinear analyses require additional structural parameters and are outside this version's output scope.

\paragraph{Experimental coverage gaps.}
A fine-tuned detector baseline is not included. Such a comparison would isolate \emph{strong visual detector + post-processing} from \emph{MLLM agent + constraint-driven revisits} and is a direct extension of the current evaluation.

\paragraph{Cost and latency.}
Targeted revisits raise the tool-call count and token cost on harder sheets. Wall-clock and token-cost comparisons are not included. Selective revisit triggering, OCR/CV caching, and policy distillation are practical directions for reducing this overhead.

%% file: sections/ethics.tex
\paragraph{Data provenance and anonymization.}
The 300 sheets used for evaluation come from real structural construction archives released under partner permission. Project names, client identifiers, designer signatures, addresses, and similar information were removed before the system processed any sheet. The data contains no personally identifiable information.

\paragraph{Engineering safety and human oversight.}
BPA produces simulation-ready frame models that help structural engineers start their modeling and review workflows; it does not replace qualified-engineer judgment. Any application involving load-bearing modifications, seismic retrofit, code-compliance checks, or safety-critical decisions must be reviewed and signed off by an engineer-of-record.

\paragraph{Uncertainty disclosure.}
\begin{sloppypar}
The \texttt{confidence}, \texttt{status}, \texttt{evidence\_refs}, and \texttt{unresolved\_items} fields in the output JSON are designed to reduce false confidence. Downstream systems should display these fields explicitly rather than presenting only the final 3D model or FEM file.
\end{sloppypar}

\paragraph{MLLM API dependence and reproducibility.}
When a commercial MLLM API is used, reproducibility is affected by model version and service-side policy. The supplementary material provides the model version, prompts, tool configuration, intermediate simulatable 3D model, and evidence trace, so that results can be re-evaluated on different backbones.

\paragraph{AI-assistant disclosure.}
The authors used LLMs to assist with English phrasing, code drafting, and literature screening; all technical claims, experimental numbers, and citations were independently verified by the authors.

%% file: sections/acknowledgments.tex
This work was supported by the Undergraduate ``Qiyan'' Program of Beijing Natural Science
Foundation (Grant No.~QY26402) and the Beijing Nova Program (Grant No.~202604841290).

%% file: sections/appendix.tex
\section{Dataset statistics}
\label{sec:appendix_dataset}

\begin{table}[H]
\centering
\small
\begin{tabular*}{\columnwidth}{@{\extracolsep{\fill}}lrl@{}}
\toprule
\textbf{Tier} & \textbf{Projects} & \textbf{Description} \\
\midrule
Simple  & 10 & 2--3 floors, regular grid \\
Medium  & 7  & 4--6 floors, mild interference \\
Complex & 3  & 7+ floors, composite grid \\
\textbf{Total} & \textbf{20} & \textbf{300 sheets, 102 floors} \\
\bottomrule
\end{tabular*}
\caption{Dataset statistics by difficulty tier: 300 sheets and 102 floors from 20 projects.}
\label{tab:dataset}
\end{table}

\section{OCR multi-scale protocol}
\label{sec:appendix_ocr}

The OCR subsystem returns a multi-scale candidate set with confidence scores for each bounding box specified by the MLLM. The MLLM then selects the most consistent reading under the surrounding axis grid, engineering-annotation conventions, and the current constraint context, thereby preventing OCR top-1 errors from propagating into the simulatable 3D model.

\section{Metric definitions}
\label{sec:appendix_metrics}

\paragraph{Axis F1.}
Matches when the predicted axis line is equivalent in label to the ground-truth axis line and position IoU $\geq 0.85$.

\paragraph{Column F1.}
After axis-grid matching, matches when the predicted column sits at the same grid intersection as the ground-truth column and the column ID is equivalent.

\paragraph{Beam Topology F1.}
Matches when the beam ID matches and the endpoint axis multiset matches.

\paragraph{KL Text Binding F1.}
Matches when the normalized KL text content matches and the target beam ID matches.

\paragraph{JSON Schema Validity.}
The output JSON passes the evidence-linked JSON schema with zero errors.

\paragraph{Structured-Model Completeness (SMC).}
For project $p$, let $s_p=1$ iff $C_{\textrm{3D}}$ validation is enabled, the normalized evaluator status is \texttt{OK}, the predicted column and primary-beam inventories are both nonempty, and each has at least one ground-truth match; otherwise $s_p=0$. We report $\frac{1}{20}\sum_{p=1}^{20}s_p$. Thus A4, which disables $C_{\textrm{3D}}$ validation by construction, has $s_p=0$ for every project.

\section{Per-tier results}
\label{sec:appendix_pertier}

To support the stratification observation in Section~\ref{sec:experiment} (constraint contribution stratifies with complexity), Tables~\ref{tab:d1} and~\ref{tab:d2} report the macro-averaged Beam F1 and Col F1 for B5 and the key ablations on each difficulty tier (simple / medium / complex; project counts 10 / 7 / 3, see Table~\ref{tab:dataset}).

\begin{table}[H]
\centering
\scriptsize
\setlength{\tabcolsep}{3pt}
\begin{tabular*}{\columnwidth}{@{\extracolsep{\fill}}lcccc@{}}
\toprule
\textbf{Variant} & \textbf{Simple} & \textbf{Medium} & \textbf{Complex} & \textbf{Overall} \\
 & (10) & (7) & (3) & (20) \\
\midrule
B5 Full BPA            & 1.000 & 1.000 & 0.960 & 0.994 \\
A3: no x-sheet aggr.   & 0.662 & 0.541 & 0.291 & 0.564 \\
A4: no $C_{\textrm{3D}}$ closure & 0.653 & 0.507 & 0.314 & 0.551 \\
A6: no axis adjud.\    & 0.571 & 0.217 & 0.050 & 0.369 \\
\bottomrule
\end{tabular*}
\caption{Per-tier Beam F1.}
\label{tab:d1}
\end{table}

\begin{table}[H]
\centering
\scriptsize
\setlength{\tabcolsep}{3pt}
\begin{tabular*}{\columnwidth}{@{\extracolsep{\fill}}lcccc@{}}
\toprule
\textbf{Variant} & \textbf{Simple} & \textbf{Medium} & \textbf{Complex} & \textbf{Overall} \\
 & (10) & (7) & (3) & (20) \\
\midrule
B5 Full BPA            & 1.000 & 1.000 & 1.000 & 1.000 \\
A3: no x-sheet aggr.   & 0.689 & 0.402 & 0.205 & 0.516 \\
A4: no $C_{\textrm{3D}}$ closure & 0.825 & 0.552 & 0.282 & 0.648 \\
A6: no axis adjud.\    & 0.849 & 0.497 & 0.104 & 0.614 \\
\bottomrule
\end{tabular*}
\caption{Per-tier Col F1.}
\label{tab:d2}
\end{table}

\section{Terminology}
\label{sec:appendix_terminology}

Table~\ref{tab:terminology} lists the paper-specific terms together with their definitions for quick reference.

\begin{table}[H]
\centering
\small
\begin{tabular*}{\columnwidth}{@{\extracolsep{\fill}}>{\raggedright\arraybackslash}p{0.32\columnwidth}>{\raggedright\arraybackslash}p{0.58\columnwidth}@{}}
\toprule
\textbf{Term} & \textbf{Definition} \\
\midrule
sheet & A single scanned structural blueprint \\
region & A semantic region within a sheet \\
entity & A structural object in $H$ (axis / column / beam / floor / material) \\
evidence ($E$) & A localized OCR / CV / MLLM observation with pixel provenance \\
conflict ($q$) & A validator's violation report on $H$ \\
$H$ & The project-level evidence-linked structural representation \\
validator & A callable checker $v: H \to \textrm{List}[\textrm{Conflict}]$ \\
KL / KZ & Frame-beam / frame-column labeling prefixes (e.g., \texttt{KL8(4)} = frame beam 8, four-span) \\
\texttt{axis\_px\_\allowbreak span\_\allowbreak ratio} & Consistency check between axis-span pixel ratios and annotated span ratios ($\pm15$\%); ablation A5 \\
axis adjudication & The MLLM's final decision on axis labels, integrating OCR / CV evidence \\
targeted revisit & MLLM revisiting local evidence and revising an entity, triggered by a validator conflict \\
SMC & Project-level completion gate for nonempty matched column/beam inventories under $C_{\textrm{3D}}$ validation \\
\bottomrule
\end{tabular*}
\caption{Terminology used throughout the paper.}
\label{tab:terminology}
\end{table}

\section{Feature comparison with prior agents}
\label{sec:appendix_rw}

Table~\ref{tab:rw} contrasts BPA with representative verification, reflection, and tool-augmented agents along four design dimensions.

\begin{table}[H]
\centering
\scriptsize
\setlength{\tabcolsep}{3pt}
\begin{tabular*}{\columnwidth}{@{\extracolsep{\fill}}lllll@{}}
\toprule
\textbf{Work} & \textbf{Trigger} & \textbf{Signal} & \textbf{Granularity} & \textbf{Sym.} \\
\midrule
ReAct            & model            & NL thought        & task     & no \\
Reflexion        & model            & NL self-critique  & episode  & no \\
Self-Refine      & model            & NL critique       & response & no \\
Self-Consist.\   & sampling         & majority vote     & answer   & no \\
CRITIC           & tool             & tool output       & response & part. \\
PRM              & external         & step score        & step     & no \\
\textbf{BPA}     & \textbf{schema}  & \textbf{JSON conf.} & \textbf{entity} & \textbf{yes} \\
\bottomrule
\end{tabular*}
\caption{Feature comparison with prior work. BPA differs in trigger source (external schema validator), signal type (structured JSON conflict), revisit granularity (entity-level), and explicit evidence-grounded revisions.}
\label{tab:rw}
\end{table}

\section{Representative cases}
\label{sec:appendix_cases}

Table~\ref{tab:cases} summarizes three representative cases by failure mode, triggering constraint, and revisit outcome; Section~\ref{sec:cases} discusses Case~2 (the A6 counterexample) in detail.

\begin{table}[H]
\centering
\small
\begin{tabular*}{\columnwidth}{@{\extracolsep{\fill}}>{\raggedright\arraybackslash}p{0.10\columnwidth}>{\raggedright\arraybackslash}p{0.30\columnwidth}>{\raggedright\arraybackslash}p{0.20\columnwidth}>{\raggedright\arraybackslash}p{0.22\columnwidth}@{}}
\toprule
\textbf{Case} & \textbf{Failure mode} & \textbf{Trigger} & \textbf{Outcome} \\
\midrule
1 & CV axis misreport from a dimension annotation & $C_{\textrm{axis}}$ span-ratio mismatch & MLLM corrects misreport; entities consistent \\
2 (A6) & CV accepted $\to$ ghost axis & none (gate off) & Errors propagate; F1 collapses \\
3 & $C_{\textrm{3D}}$ passes; OpenSees runs with anomalous top-floor displacement & semantic constraints exhausted & Qualified-engineer review \\
\bottomrule
\end{tabular*}
\caption{Representative cases by failure mode, triggering constraint, and revisit outcome.}
\label{tab:cases}
\end{table}

\section{Adjudicated consensus error profile}
\label{sec:appendix_errors}

Table~\ref{tab:errors} reports the four-category adjudicated consensus breakdown of residual B5 errors over the 300-sheet evaluation; annotations were carried out independently by two structural engineers, with disagreements arbitrated by a third.

\begin{table}[H]
\centering
\small
\begin{tabular*}{\columnwidth}{@{\extracolsep{\fill}}p{0.24\columnwidth}r p{0.50\columnwidth}@{}}
\toprule
\textbf{Category} & \textbf{Share} & \textbf{Trigger} \\
\midrule
Perception   & 45\% & Small font, low contrast, rotated, or aged scans \\
Topology     & 25\% & Adjacent spans share a column but differ in section; beam-segment merge errors \\
Binding      & 20\% & Densely placed KL labels may bind to a neighboring beam \\
Completeness & 10\% & Critical notes lie on sheets outside the current working set \\
\bottomrule
\end{tabular*}
\caption{Adjudicated consensus error profile for B5 over the 300-sheet evaluation.}
\label{tab:errors}
\end{table}

\section{BPA main loop pseudocode}
\label{sec:appendix_algorithm}

\begin{algorithm}[H]
\caption{BPA main loop. Outer loop walks through five sequential stages; inner loop runs a seven-step assess-to-revisit cycle until conflicts are resolved or the stage budget is exhausted. Validators are aggregated as $\textsc{validate}(H, C) = \bigcup_{v \in C} v(H)$, each $v: H \to \textrm{List}[\textrm{Conflict}]$.}
\label{alg:bpa}
\begin{algorithmic}[1]
\Require project $P$, MLLM backbone $M$, constraints $C = \{C_{\textrm{axis}}, C_{\textrm{geo}}, C_{\textrm{anno}}, C_{\textrm{3D}}\}$
\Ensure simulatable 3D model $H$, simulation-ready file $F$
\State $H \gets \textsc{init\_H}(P)$
\State $\textit{cache} \gets \textsc{ocr\_cache}(P)$
\State $\textit{regions} \gets \textsc{analyze\_regions}(P)$
\For{stage $\in$ \{axes, columns, beams, validation, 3D-check\}}
  \Repeat
    \State $a \gets M.\textsc{assess}(H, \textrm{stage})$
    \State $(r, t) \gets M.\textsc{plan}(H, \textrm{stage}, a)$
    \State $o \gets \textsc{execute}(t, r)$
    \State $e \gets M.\textsc{interpret}(o)$
    \State $H \gets M.\textsc{update}(H, e)$
    \State $Q \gets \textsc{validate}(H, C)$
    \If{$Q = \emptyset$} \textbf{break} \EndIf
    \State $H \gets M.\textsc{revisit}(H, Q)$
  \Until{stage budget exhausted}
\EndFor
\State $F \gets \textsc{sim\_export}(H)$
\State \Return $H, F$
\end{algorithmic}
\end{algorithm}